\documentclass[lettersize,journal]{IEEEtran}
\usepackage{amsmath,amsfonts}
\usepackage{algorithmic}
\usepackage{algorithm}
\usepackage{array}
\usepackage[caption=false,font=normalsize,labelfont=sf,textfont=sf]{subfig}
\usepackage{textcomp}
\usepackage{stfloats}
\usepackage{url}
\usepackage{verbatim}
\usepackage{graphicx}
\usepackage{hyperref}
\usepackage{cite}
\usepackage{multirow}
\usepackage{threeparttable}
\usepackage{booktabs}
\usepackage{color} 
\usepackage{amssymb}
\usepackage{amsmath}
\usepackage{float}
\begin{document}

\title{Wavelet-based Decoupling Framework for low-light Stereo Image Enhancement}

\author{\IEEEauthorblockN{Shuangli Du\textsuperscript{1},Siming Yan\textsuperscript{1},Zhenghao Shi\textsuperscript{1,\dag},Zhenzhen You\textsuperscript{1},Lu Sun\textsuperscript{1}}
	\\
	\IEEEauthorblockA{\textit{1. Shaanxi Key Laboratory for Network Computing and Security Technology, School of Computer Science and Engineering, Xi'an University
			of Technology, Xi'an, China}}
	\\
	\IEEEauthorblockA{\dag	Correspondence:\href{mailto:ylshi@xaut.edu.cn}{ylshi@xaut.edu.cn}}

}

\maketitle

\begin{abstract}
	Low-light images suffer from complex degradation, and existing enhancement methods often encode all degradation factors within a single latent space. This leads to highly entangled features and strong black-box characteristics, making the model prone to shortcut learning. To mitigate the above issues, this paper proposes a wavelet-based low-light stereo image enhancement method with feature space decoupling. Our insight comes from the following findings: (1) Wavelet transform enables the independent processing of low-frequency and high-frequency information. (2) Illumination adjustment can be achieved by adjusting the low-frequency component of a low-light image, extracted through multi-level wavelet decomposition. Thus, by using wavelet transform the feature space is decomposed into a low-frequency branch for illumination adjustment and multiple high-frequency branches for texture enhancement. Additionally, stereo low-light image enhancement can extract useful cues from another view to improve enhancement. To this end, we propose a novel high-frequency guided cross-view interaction module (HF-CIM) that operates within high-frequency branches rather than across the entire feature space, effectively extracting valuable image details from the other view. Furthermore, to enhance the high-frequency information, a detail and texture enhancement module (DTEM) is proposed based on cross-attention mechanism. The model is trained on a dataset consisting of images with uniform illumination and images with non-uniform illumination. Experimental results on both real and synthetic images indicate that our algorithm offers significant advantages in light adjustment while effectively recovering high-frequency information. The code and dataset are publicly available at: https://github.com/Cherisherr/WDCI-Net.git.
\end{abstract}


\section{Introduction}
\IEEEPARstart{U}{nlike} monocular enhancement, binocular stereo image enhancement can improve image quality by leveraging the complementarity and redundancy of the two views \cite{1}. Existing algorithms typically encode image content, light distribution, and noise into a latent space, then perform information exchange within that space, and finally decode the feature into the enhanced result \cite{2}.  However, low-light images suffer from complex degradation, and regarding all the degradation factors as a whole will result in highly entangled features and strong black-box characteristic, making the model prone to shortcut learning and ultimately reducing its generalization ability.

Recently, for monocular image enhancement, several decoupling algorithms have been proposed  to disentangle the degradation factors and process them separately, including approaches based on Retinex decomposition \cite{3}, \cite{4}, \cite{5}, \cite{6}, \cite{7} and Fourier transform \cite{8}, \cite{9}, \cite{10}. Retinex-based methods typically require a well-designed decomposition algorithm to separate a given image into illumination and reflectance components. This transforms the original single-image mapping problem into two subproblems: illumination mapping and reflectance mapping \cite{11}. However, developing an efficient and stable Retinex decomposition framework remains highly challenging. Fourier-based approaches often divide the feature space into the amplitude and the phase component, where the amplitude primarily captures illumination information, while the phase encodes geometric texture details \cite{12}. However, noise affects both amplitude and phase, making feature disentanglement even more challenging. Additionally, while convolution operations in the frequency domain can capture long-range contextual information, they may disrupt local image structures in the spatial domain, unlike Transformer-based methods. We observe that ripple artifacts emerge when local information is amplified.

\begin{figure*}[t]\centering
	\includegraphics[width=6.5in]{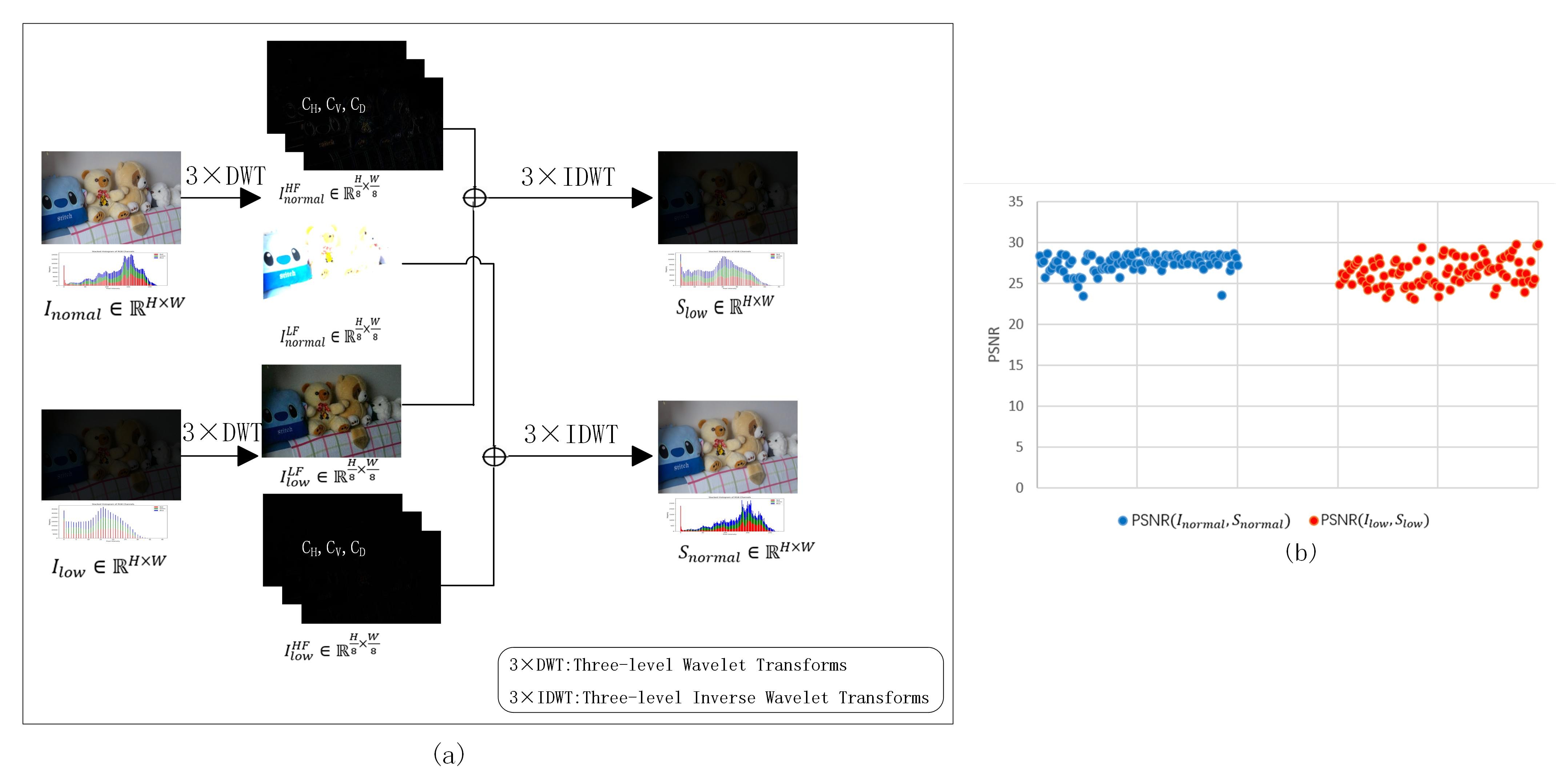}
	\caption {(a) Examples of image reconstruction via exchange the low-frequency components; (b) Illustration of PSNR ($I_{normal}$, $S_{normal}$) and PSNR ($I_{low}$, $S_{low}$).}
	\label{pho:a}
\end{figure*}

Inspired by the work \cite{13}, this paper proposes a novel method for stereo image enhancement based on wavelet transform. Different from Fourier transform, Wavelet transform can provide frequency information on different scales, and is effective to capture local features and carry out finer analysis. For a given image $I$, wavelet transform can decompose it into four frequency subbands:
\begin{equation}\label{eq0}
	{{cA,cH,cV,cD}} = DWT(I)
\end{equation}where $DWT$ represents the Discrete Wavelet Transform. $cA$ denotes the low-frequency components. $cH$, $cV$ and $cD$ represent the high-frequency components in the horizontal, vertical and diagonal directions, respectively. The low-frequency component $cA$ can be further processed using wavelet transform to achieve a second-level decomposition. We select 100 pairs of low-light/normal-light images ($I_{low}$, $I_{normal}$) from LOL dataset \cite{49}. Each image pair is decomposed with three-level wavelet transform, and the low-frequency components obtained are exchanged for image reconstruction, as illustrated in Figure \ref{pho:a}(a). Experimental results show that the low-frequency component has a significant impact on the reconstruction results. Specifically, combining the low-frequency component from $I_{normal}$ with the high-frequency components from $I_{low}$ with inverse wavelet transform, the reconstructed image $S_{normal}$ is close to $I_{normal}$. Conversely, using the low-frequency component from $I_{low}$ and the high-frequency components from $I_{normal}$, the reconstructed image $S_{low}$ is similar to $I_{low}$. This finding is also supported by the PSNR (peak signal-to-noise ratio) values between $S_{normal}$ and $I_{normal}$, as well as between $S_{low}$ and $I_{low}$, as shown in Figure \ref{pho:a}(b).

The experiments show that illumination adjustment can be achieved by adjusting the low-frequency component of low-light images, extracted through multi-level wavelet decomposition. This motivate us to adjust illumination and recover image details independently, thus this paper proposes a wavelet-based method with feature space decoupling. By using wavelet transform the feature space is decomposed into a low-frequency branch for illumination adjustment and multiple high-frequency branches for texture enhancement. We observe that illumination adjustment can be handled with single-view information, while the high-frequency branches require cross-view interaction to recover fine details. Accordingly, we propose a novel high-frequency guided cross-view interaction module (HF-CIM) based on parallax attention mechanism (PAM). Furthermore, a detail and texture enhancement module (DTEM) is proposed based on cross-attention mechanism to strengthen the high-frequency components. Experimental results demonstrate the advantages of the proposed method in image enhancement and generalization capability. In summary, the main contributions of this work are as follows:

(1)  A novel wavelet-based decoupling stereo image enhancement method is proposed. This method processes image illumination and textures details separately, effectively improving brightness, contrast adjustment and details recovery.

(2)  A novel cross-view feature interaction module HF-CIM based on PAM is proposed. To suppress noise influence, high-frequency features in vertical, horizontal and diagonal directions are first fused, and parallax attention is then estimated from the fused features to enable cross-view interaction.

(3)  A new high-frequency information enhancement module DTEM is proposed.  To fully exploit the correlations between high-frequency features in different directions, we first fuse them into a complete feature map, and then use it to guide enhancement.

(4)  To enhance model robustness, the training dataset includes low-light images with both uniform and non-uniform illumination. Experiments demonstrate that the trained model can effectively handle images captured in complex and dynamic low-light conditions.

\section{Related Work}

In terms of methodology, low-light image enhancement approaches can be classified into traditional methods and deep learning-based methods. In terms of the target data, low-light image enhancement methods can be divided into single-image enhancement, stereo-image enhancement, and video enhancement. In the following, we introduce deep learning-based methods for single-image and stereo-image enhancement.

\subsection{Low-light Single-image Enhancement}

\noindent \textbf{Non-decoupling Method:} Non-decoupling methods generally adopt a holistic end-to-end network to enhance images without explicitly dividing the enhancement into sub-tasks. However, these methods often encounter challenges such as severe feature entanglement and limited interpretability. LLNet \cite{21} is the first work introducing deep learning to low-light image enhancement. Subsequently, many end-to-end networks have been proposed, focusing on efficient network design. Ren et al. \cite{22} introduced a hybrid network combining a content recovery flow and an edge detail enhancement flow. Lim and Kim \cite{23} proposed a deep-stacked Laplacian restorer that adjusts brightness and recovers local details at each pyramid level, progressively merging them in the image space. Li et al. \cite{24} designed a multi-stage enhancement where each stage takes an attention map and the previous output as inputs. Later, to overcome the challenge of collecting low-light/normal-light image pairs for supervised training, several un-supervised algorithms have been proposed. EnlightenGAN \cite{51} utilized generative adversarial networks to generate enhanced images that exhibit the statistical properties of normal-light images. Zero-DCE \cite{46}, a zero-reference method, learns pixel-wise high-order curves to dynamically adjust each pixel value, thereby achieving brightness enhancement. Its lightweight architecture makes it suitable for real-time low-light image processing.

\noindent \textbf{Decoupling Method:} The existing decoupling methods often processing illumination and texture separately, thereby improving detail restoration and illumination correction. They typically leverage methods such as the Retinex model, Fourier transform, or wavelet transform to decompose the original single supervision task into two or more independent subtasks. Retinex-based methods \cite{11}, \cite{38}, \cite{39} typically consist of three subnetworks dedicated to decomposition, reflectance recovery, and illumination adjustment, respectively. Due to the lack of ground-truth decomposition results for reflectance and illumination, designing appropriate constraint for decomposition is particularly important. Common constraints include enforcing consistent reflectance across paired images and ensuring smooth illumination \cite{40}. Recently, Jiang at al. \cite{50} proposed a content-transfer decomposition network that performs Retinex decomposition in the latent space instead of image space, and then they used diffusion model to generate enhanced result. 

  Another commonly used decoupling strategy involves separating an image into the low- and high-frequency information and processing them individually. For example, Yang et al. \cite{42} jointly learn the mapping functions for the low-frequency component, high-frequency component, and the overall image using paired data. In Fourier domain, lightness degradation primarily affects the amplitude component, whereas other degradations are mostly concentrated in the phase component. Based on this observation, Li et al. \cite{10} proposed an enhancement method for ultra-high-definition (UHD) images. They first adjust the phase and amplitude in the low-resolution domain and then apply a lightweight super-resolution process. However, we found noise in images will affect both amplitude and phase, thereby increasing the difficulty of feature disentanglement. Differently, the wavelet transform not only separates the image content from the noise, but also allows downsampling without loss of information. Zou et al. \cite{13} combined wavelet transform with state-space model to handle UHD images. Wavelet transform, and channel-wise Mamba modules are integrated by Tan et al. \cite{52} to enhance image brightness and details across multiple scales. However, currently, no decoupling methods have been proposed for stereo image enhancement, and this paper is the first attempt in this area.

\subsection{Stereo Image Restoration and Enhancement}

\begin{figure*}[t]\centering
	\includegraphics[width=6in]{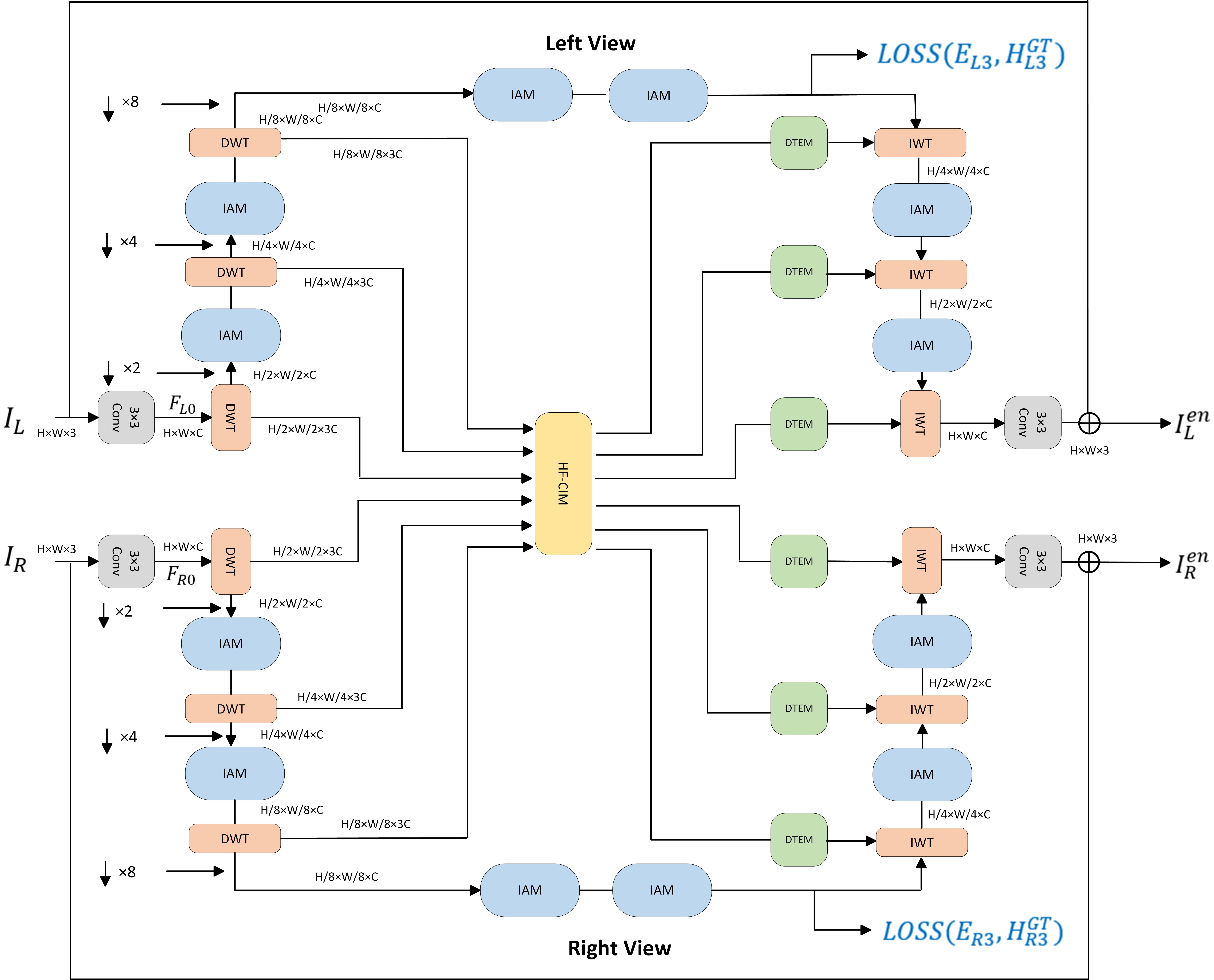}
	\caption{The overall architecture of WDCI-Net includes two weight-shared branches for processing the left and right views, respectively. The method comprises three main modules, namely HF-CIM, DTEM, and IAM. HF-CIM is used for multi-scale cross-view interaction of high-frequency information, DTEM further enhances high-frequency features and suppresses noise, and IAM processes low-frequency information and recovers illumination.}
	\label{pho:3}
\end{figure*}

\begin{figure}[!t]\centering
	\includegraphics[width=3.2in]{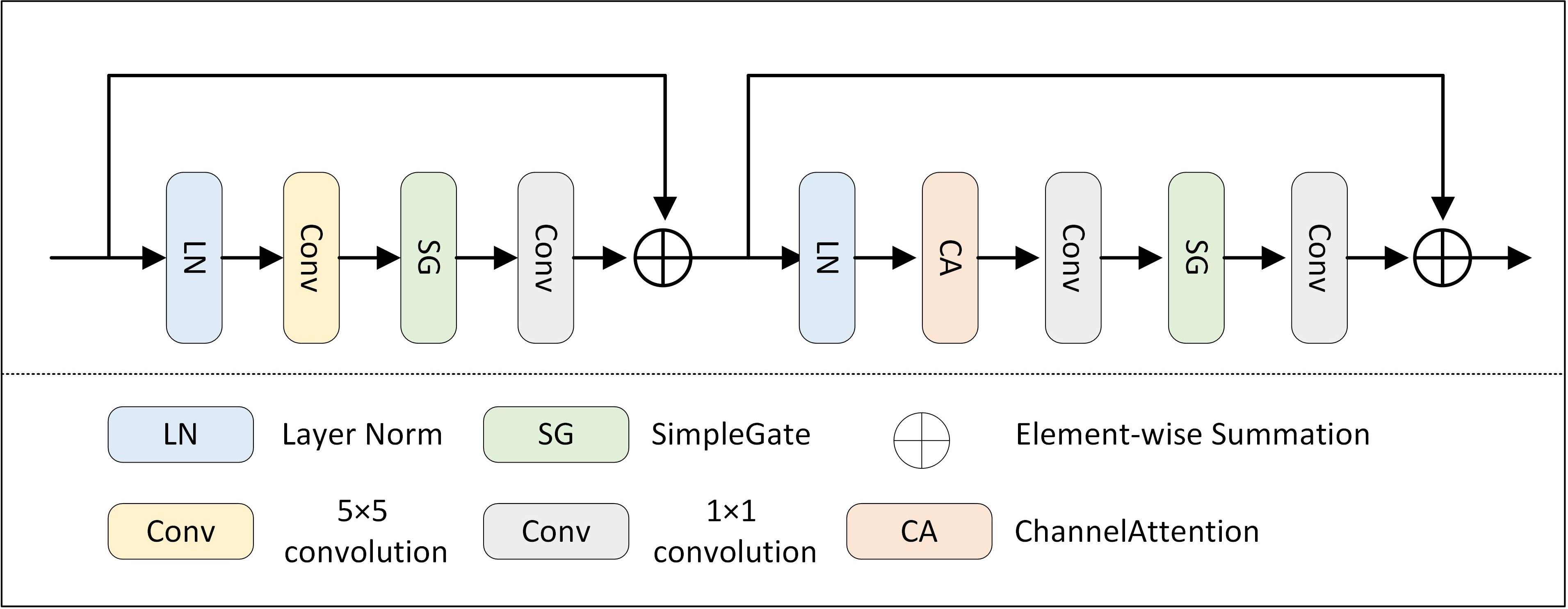}
	\caption{The detailed process of the the Illumination Adjustment Module (IAM).}
	\label{pho:IAM}
\end{figure}

Different from single image processing, stereo image restoration pays more attention on mining complementary information between the left and right views. Jeon et al. \cite{31} directly used parallax prior information to improve the spatial resolution of stereoscopic images. Recently, Wang et al. \cite{32} proposed an unsupervised parallax attention mechanism (PAM) to learn stereo correspondence, which has driven the development of stereo image restoration. Baed on PAM, Zhu et al. \cite{33} used cross attention to fuse features of left and right views without need of disparity prior measure stage. Guo et al. \cite{34} integrated PAM with Transformer architecture, introducing a novel Transformer-based parallax fusion module. Cheng et al. \cite{35} designed a hybrid network that leverages a converter-based model for single-image enhancement and a CNN-based model, equipped with PAM, for stereo information fusion. 

In 2022, Huang et al. \cite{36} first applied PAM in the field of stereo low-light image enhancement. They performed PAM on low-resolution features, followed by up-sampling to infer high-resolution feature interactions. To enhance the reliability of cross-view feature interaction, Zheng et al. \cite{2} applied PAM across multi-scale features. To mitigate noise interference, Zhao et al. \cite{1} conducted cross-view interaction within an image space with low-frequency information ehanced. In contrast, Zhang et al. \cite{37} utilized large convolutional kernels to capture cross-view interactions at multiple scales. Different from the above methods, this paper performs cross-view feature interaction only on the high-frequency information rather than the entire feature space.

\section{Method}

\subsection{Overall Architecture}
The overall framework of the proposed wavelet-based decoupling cross-view interaction network WDCI-Net for stereo image enhancement is illustrated in Figure \ref{pho:3}. It consists of five module: wavelet transform and downsampling, IAM for illumination adjustment, HF-CIM for high-frequency feature interaction, DTEM for high-frequency information enhancement, and inverse wavelet transform for image reconstruction. Unlike information-lossy downsampling, we use wavelet transform, which not only separates features into low-frequency and high-frequency components but also achieves down-sampling without any information loss. 

Specifically, let's take the left view as an example, given a low-light image $I_{L}\in\mathbb{R}^{H\times W\times3}$, We first apply 3×3 convolutions to obtain low-level embedding features $F_{L0}\in\mathbb{R}^{H\times W\times C}$, where $H$, $W$ and $C$ represent the height, width and number of channels respectively. Subsequently, three-level DWT is applied to the initial feature map $F_{L0}$, decomposing it into low- and high-frequency components across multiple scales. Specifically, each DWT yields a low-frequency feature map $F_{Li}(i=1,2,3)$, along with three high-frequency feature maps $(V_{Li},H_{Li},D_{Li})$, capturing vertical, horizontal, and diagonal detail information, respectively. These feature maps are downsampled to spatial resolutions of  $\frac{H}{2}\times\frac{W}{2},$$\frac{H}{4}\times\frac{W}{4}$ and $\frac{H}{8}\times\frac{W}{8}$ respectively. To fully exploit the useful information in the original image $I_{L}\in\mathbb{R}^{H\times W\times3}$, we perform independent multi-scale downsampling. Specifically, we adopt a combination of PixelUnshuffle and convolution: PixelUnshuffle reduces spatial resolution while increasing the number of channels, followed by 1×1 convolution to compress the channels, producing a feature map $F_{\downarrow\times x}(x=2,4,8)$ that matches the channel dimension of the low-frequency feature map $F_{Li}$. Then, $F_{Li}$ and the corresponding $F_{\downarrow\times x}$ are concatenated along the channel dimension and fused via convolution to construct a more informative feature representation. 

Our network is composed of one low-frequency branch and three high-frequency branches. The low-frequency branch performs feature extraction and illumination restoration through IAM. Inspired by multi-task learning, we impose supervisory constraints on the low-frequency branch to make it focus on illumination and color adjustments. Similarly, the right-view image is processed in the same manner as the left-view image. For the high-frequency layer, the features from the left and right views are first fused by HF-CIM to facilitate cross-view information interaction. And then, the output features are further refined by DTEM to enhance image details and effectively suppress noise. Finally, the restored high- and low-frequency information is progressively reconstructed through inverse wavelet transforms, producing the enhanced results ${I}_{L}^{en}$ and ${I}_{R}^{en}$.

\begin{figure*}[t!]\centering
	\includegraphics[width=5.5in]{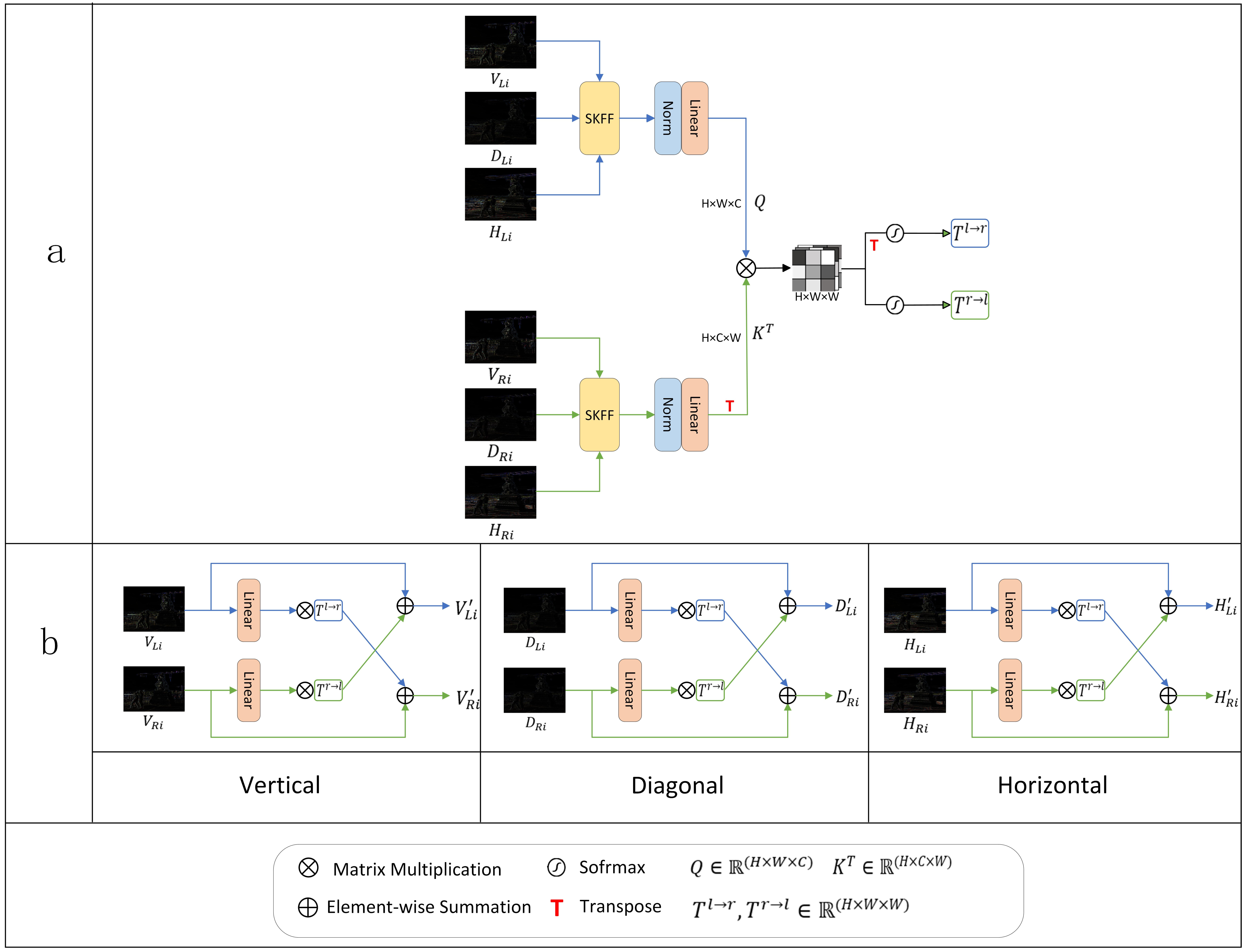}
	\caption{(a) The first stage of the HF-SMM module: computing and generating the attention map; (b) The second stage: using the attention map to produce the interacted high-frequency features.}
	\label{pho:4}
\end{figure*}

\subsection{Illumination Adjustment Module (IAM)}

As shown in Figure \ref{pho:IAM}, IAM is employed for feature extraction and illumination restoration. This module consists of two stages, each utilizing a residual structure. In the first stage, large-kernel convolutions are adopted to effectively capture long-range dependencies, leveraging their larger receptive field to extract global illumination information from the image. While this stage primarily focuses on spatial information, it pays less attention to channel-wise features. Therefore, the second stage applies a channel attention mechanism, which adaptively adjusts the importance of each channel, highlighting features closely related to illumination restoration and enhancing the representation of critical illumination cues. Both stages integrate the SimpleGate (SG) unit to adaptively amplify key information and suppress redundant data, thereby improving the representational capacity.

For the low-frequency branch, we leverage the concept of multi-task learning by treating the restoration of low-frequency components as a sub-supervised task. To this end, we design three complementary loss functions—frequency domain loss, spatial domain loss, and VGG perceptual loss—to jointly constrain this process, ensuring that the feature encoding in this module is oriented toward the objectives of illumination and color adjustments.

\subsection{High-Frequency Guided Cross-view Interaction Module (HF-CIM)}

The details of the proposed HF-CIM are illustrated in Figure \ref{pho:4}. HF-CIM module is based on PAM. PAM searches for corresponding features in the left view within a predefined disparity range and compute feature similarity between two views to generate one attention map, which guides the weighted fusion of cross-view features. Specifically, the HF-CIM consists of two stages. The first stage is illustrated in Figure \ref{pho:4}(a), for the source view, we first employ the Selective Kernel Feature Fusion (SKFF) \cite{55} module to integrate high-frequency information in the vertical, diagonal, and horizontal directions. On one hand, this integration significantly reduces the computational overhead that would result from processing each directional feature independently; on the other hand, it helps suppress noise commonly present in high-frequency features, thereby improving the stability and robustness of the feature representation. The fused left and right views are then projected to obtain the view-specific feature projection matrices ${Q}\in\mathbb{R}^{H\times W\times C}$ and $K^T\in\mathbb{R}^{(H\times C\times W)}$. Finally, we compute the feature correlation between arbitrary positions along the epipolar line by performing set-aware matrix multiplication between $Q$ and $K^T$, thereby generating the disparity attention maps $T^{r\rightarrow l}$ and $T^{l\rightarrow r}$. $\mathrm{T^{l\to r}}$ is the attention map from the left view features to the right view features, and $\mathrm{T^{r\to l}}$ is the attention map from the right view features to the left view features. The second stage is illustrated in Figure \ref{pho:4}(b), taking $\mathrm{T^{l\to r}}$ as an example, after obtaining the feature attention map, we perform matrix multiplication between $\mathrm{T^{l\to r}}$ and the high-frequency features of the left view, and then add the results to the corresponding high-frequency features of the right view. Finally, the fused features of the right view in the three directions are obtained. The features of the left view are similarly obtained using this method. The formal representation is as follows:

\begin{figure*}[t!]\centering
	\includegraphics[width=5.5in]{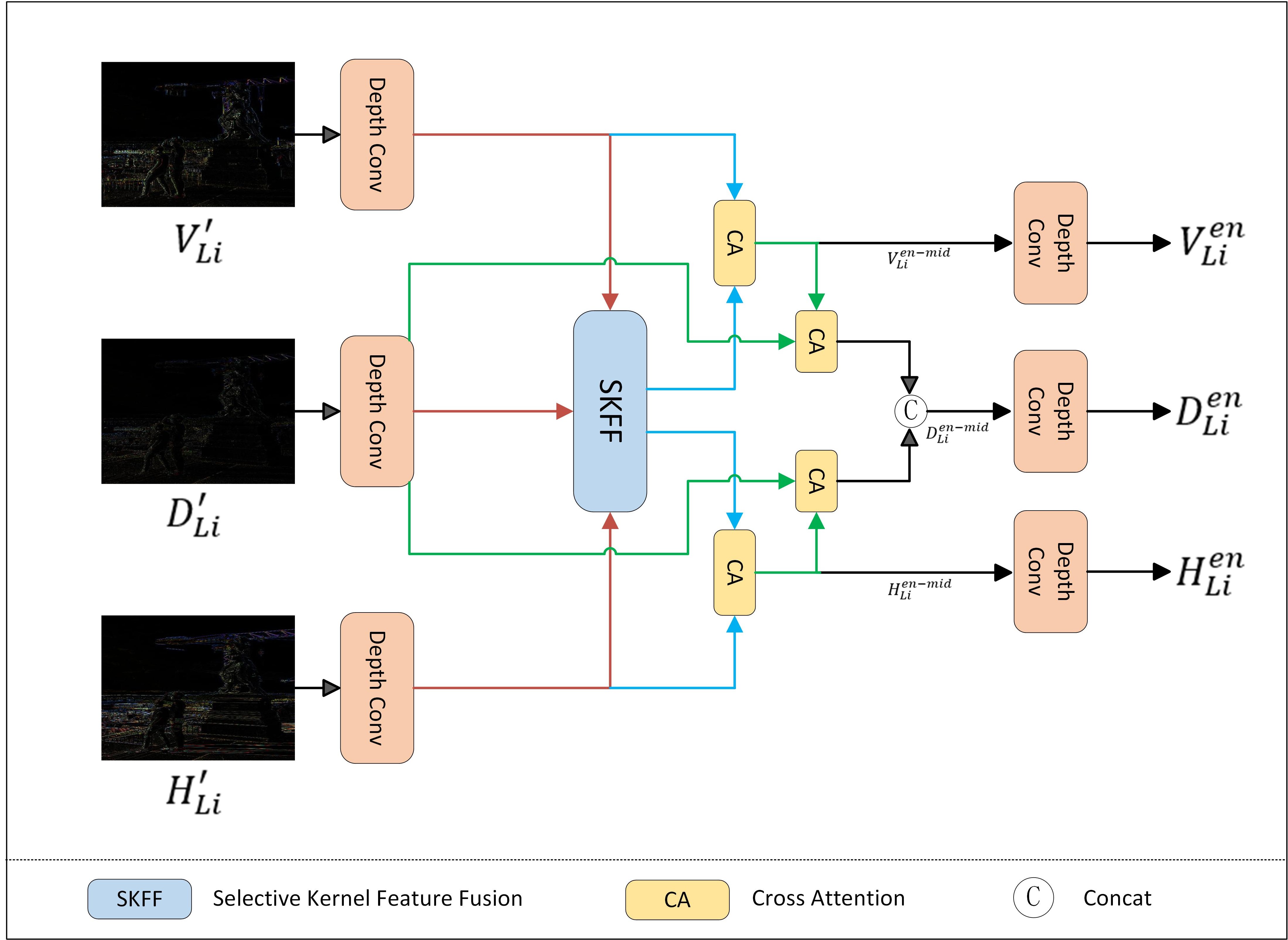}
	\caption{The detailed process of the the Detail and Texture Enhancement Module (DTEM).}
	\label{pho:5}
\end{figure*}

\begin{equation}
	\label{eq1}
	\left\{
	\begin{aligned}
		& {Q=linear(Norm(SKFF(V_{Li},D_{Li},H_{Li})))} 
		\\
		& {K=linear(Norm(SKFF(V_{Ri},D_{Ri,}H_{Ri})))}
		\\
		&{T^{l\to r}=softmax(K\otimes Q^{T})}
		\\
		&{T^{r\to l}=softmax(Q\otimes K^{T})}
		\\
		&{V_{Ri}^{\prime}=V_{Ri}\oplus(V_{Li}\otimes T^{l\to r})}
		\\
		&{D_{Ri}^{\prime}=D_{Ri}\oplus(D_{Li}\otimes T^{l\to r})}
		\\
		&{H_{Ri}^{\prime}=H_{Ri}\oplus(H_{Li}\otimes T^{l\to r})}
		\\
		&{V_{Li}^{\prime}=V_{Li}\oplus(V_{Ri}\otimes T^{r\to l})}
		\\
		&{D_{Li}^{\prime}=D_{Li}\oplus(D_{Ri}\otimes T^{r\to l})}
		\\
		&{H_{Li}^{\prime}=H_{Li}\oplus(H_{Ri}\otimes T^{r\to l})}
	\end{aligned}
	\right.
\end{equation}where ${V}_{{Li}},$ ${D}_{{Li}},$ ${H}_{{Li}},$ ${V}_{{Ri}},$ ${D}_{{Ri}}$ and ${H}_{{Ri}},$ represent the high-frequency features of the left and right images at the $i-th$ layer $(i = 1, 2, 3)$ in the vertical, diagonal, and horizontal directions, respectively. ${V}_{Li}^{\prime},$ ${D}_{{Li}}^{\prime},$ ${H}_{{Li}}^{\prime},$ ${V}_{{Ri}}^{\prime},$ ${D}_{{Ri}}^{\prime}$ and ${H}_{{Ri}}^{\prime}$, represent the high-frequency information after interaction. $Norm$ and $Linear$ denote the normalization and linear layers, respectively. $\otimes$ indicates matrix multiplication and $\oplus$ indicates element-wise summation.

\subsection{Detail and Texture Enhancement Module (DTEM)}

The proposed DTEM module is illustrated in Figure \ref{pho:5}. Before performing image reconstruction via inverse wavelet transform, high-frequency components need to be enhanced while suppressing noise within their spectral bands.

Specifically, taking the left view as an example, depthwise separable convolutions are first applied to the high-frequency information to extract relevant features. Then, the SKFF module is used to adaptively fuse the features from all three directions to obtain a complete feature map. Guided by the complete feature map, cross-attention mechanism is employed to first enhance the detail information in the vertical and horizontal directions, producing intermediate features $V_{Li}^{en-mid}$ and $H_{Li}^{en-mid}$. They are further used to supplement the diagonal information through cross-attention mechanism, producing feature $D_{Li}^{en-mid}$. Finally, depthwise separable convolutions are applied to $V_{Li}^{en-mid}$, $H_{Li}^{en-mid}$ and $D_{Li}^{en-mid}$, respectively, to reconstruct the high-frequency features, yielding the final enhanced outputs $V_{Li}^{en}$, $H_{Li}^{en}$, and $D_{Li}^{en}$. The formal representation is as follows:
\begin{equation}
	\label{eq2}
	\left\{
	\begin{aligned}
		&{S_{Li}=SKFF(DC(V_{Li}^{\prime},H_{Li}^{\prime},D_{Li}^{\prime}))}
		\\
		&{V_{Li}^{en-mid}=CA(S_{Li},V_{Li})}
		\\
		&{H_{Li}^{en-mid}=CA(S_{Li},H_{Li})}
		\\
		&	\begin{split}
        D_{Li}^{en-mid} &= CA(V_{Li}^{en-mid}, D_{Li}) \\
                          &\oplus CA(H_{Li}^{en-mid}, D_{Li})
      		\end{split} 
		\\
		&{V_{Li}^{en}=DC(V_{Li}^{en-mid})}
		\\
		&{H_{Li}^{en}=DC(H_{Li}^{en-mid})}
		\\
		&{D_{Li}^{en}=DC(D_{Li}^{en-mid})}
	\end{aligned}
	\right.
\end{equation}where $S_{Li}$ represents the features obtained by fusing high-frequency information from different directions through the SKFF module. $DC$ denotes the depth-wise separable convolution, $CA$ epresents the cross-attention layer, and $\oplus$ indicates element-wise addition.

\subsection{Loss Functions}

The total loss function consists of five components: the frequency domain loss $\mathcal{L}_{fre}$ and the spatial domain loss $\mathcal{L}_{spa}$ for the final output, as well as the frequency domain loss $\mathcal{L}_\mathrm{fre}^{(1/8)}$, the spatial domain loss $\mathcal{L}_\mathrm{spa}^{(1/8)}$ and the pre-trained VGG19 network loss $\mathcal{L}_\mathrm{vgg}^{(1/8)}$ for the low-frequency branch. Here, $\frac{1}{8}$ indicates that after the third $DWT$, the image resolution is downsampled to $\frac{H}{8}$ × $\frac{W}{8}$ of the original size. The overall loss function is defined as follows:

\begin{equation}\mathcal{L}=\mathcal{L}_{{fre}}+\mathcal{L}_{{spa}}+\mathcal{L}_{{fre}}^{(1/8)}+\mathcal{L}_{{spa}}^{(1/8)}+\mathcal{L}_{{vgg}}^{(1/8)}\end{equation}

By employing fourier transform, the enhanced image (${I}_{L}^{en}$,${I}_{R}^{en}$,${E}_{L3}$,${E}_{R3}$) and the ground turth (${H}_{L}^{GT}$,${H}_{R}^{GT}$,${H}_{L3}^{GT}$,${H}_{R3}^{GT}$) are mapped into the frequency domain. $\mathcal{L}_{fre}$ and $\mathcal{L}_\mathrm{fre}^{(1/8)}$ are constructed to quantify the consistency of their frequency domain distributions. The mathematical expression is as follows:
\begin{equation}
	\begin{split}
		\mathcal{L}_{{fre}}=||{FFT}({I}_{L}^{en}),{FFT}({H}_{L}^{GT})||_1\\
		+||{FFT}({I}_{R}^{en}),{FFT}({H}_{R}^{GT})||_1
	\end{split}
\end{equation}

\begin{equation}
	\begin{split}
		\mathcal{L}_{{fre}}^{(1/8)}=\|{FFT}({E}_{L3}),{FFT}({H}_{L3}^{GT})\|_1\\
		+\|{FFT}({{E}}_{{R3}}),{FFT}({H}_{{R3}}^{GT})\|_1
	\end{split}
\end{equation}where ${I}_{L}^{en}$ and ${I}_{R}^{en}$ represent the final enhanced left and right views, respectively. ${H}_{L}^{GT}$ and ${H}_{R}^{GT}$ represent the ground-truth for the left and right views, respectively. ${E}_{L3}$ and ${E}_{R3}$ represent the outputs of the low-frequency branch for the left and right views, while ${H}_{L3}^{GT}$ and ${H}_{R3}^{GT}$ represent the low-frequency information extracted from ground truth with three-level DWT. $FFT(\cdot)$ denotes the Fast Fourier Transform, and $\left\|\cdot\right\|_1$ represents the $\mathcal{L}_{1}$ loss function.

On the other hand, considering spatial features such as contours and colors, we use structural similarity to measure the pixel-level differences between the enhanced image and the ground truth. It can be described by the following formula:

\begin{equation}
	\begin{split}
		\mathcal{L}_{{spa}}{=1-SSIM(I}_{L}^{en},{H}_{{L}}^{GT})\\
		{+1-SSIM(I}_{R}^{en},{H}_{{R}}^{GT})
	\end{split}
\end{equation}

\begin{equation}
	\begin{split}
		\mathcal{L}_{{spa}}^{(1/8)}=1-{SSIM}({E}_{{L3}},{H}_{{L3}}^{{GT}})\\
		+1-{SSIM}({E}_{{R3}},{H}_{{R3}}^{{GT}})
	\end{split}
\end{equation}where $SSIM(\cdot)$  represents the structural similarity between two images, and subtracting it from 1 yields the spatial restoration loss.

The $VGG(\cdot)$ loss function, by leveraging the feature representation capabilities of deep neural networks, is able to more effectively measure the perceptual similarity between images:

\begin{equation}
	\begin{split}
		\mathcal{L}_{{vgg}}^{(1/8)}=0.0001\times||({VGG}({E}_{L3})-{VGG}({H}_{L3}^{GT}))\\
		+({VGG}({E}_{R3})-{VGG}({H}_{R3}^{GT}))||_2
	\end{split}
\end{equation}where $VGG(\cdot)$ refers to the pre-trained VGG19 network, and $\left\|\cdot\right\|_2$ denotes the $\mathcal{L}_{2}$ loss.

\section{Experimental Results and Analysis}
In this section, we first introduce the experimental setup. Then, we will present the experimental results and provide a detailed analysis.

\begin{figure*}[t]\centering
	\includegraphics[width=5.3in]{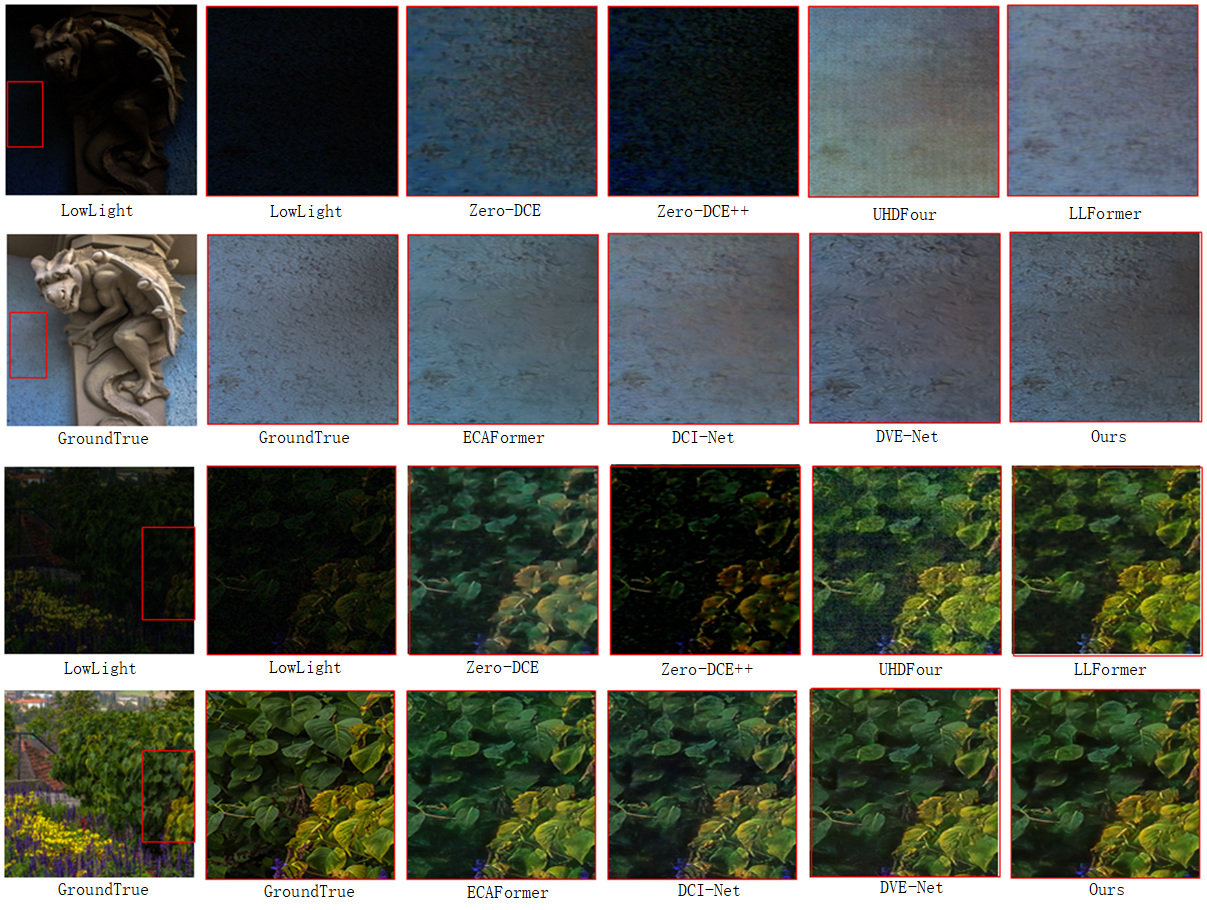}
	\caption{Visualization of the enhanced image of each method on Flickr2014 dataset. It can be seen that compared with other methods, our method effectively removes the influence of noise, at the same time, our method achieves the best details recovery.}
	\label{pho:flickr} 
\end{figure*}

\begin{table*}[t!]
	\centering
	\caption{Evaluation results of each method on  Flickr1024 and KITTI2015 datasets. The bold numbers represent the best performance. It can be observed that compared with other methods, our method shows the SOTA performance on all test datasets.}
	\begin{tablenotes}
		\item \small
	\end{tablenotes}
	\resizebox{7in}{1in}{
		\begin{tabular}{cccccccc}
			\midrule[1pt]
			&                          & \multicolumn{2}{c}{{Flickr2014}} & \multicolumn{2}{c}{{Kitti2015}} \\ \cline{3-6} 
			\multirow{-2}{*}{Methods} & \multirow{-2}{*}{Venue}                    & Left                    & Right                   & Left                      & Right                    \\
			\hline
			Zero-DCE \cite{46}                  & CVPR 2020                                     & 16.293/0.543            & 16.356/0.564            & 18.533/0.616              & 18.562/0.612             \\
			Zero-DCE++ \cite{57}                 & TPAMI2021                                      & 12.192/0.391            & 12.193/0.289            & 13.741/0.518              & 13.7930/0.515             \\
			UHDFour \cite{10}                    & ICLR 2023                                     & 19.876/0.626            & 19.857/0.626            & 24.491/0.791              & 24.119/0.786              \\
			LLFormer \cite{47}                    & AAAI 2023                                     & 24.594/0.763            & 24.630/0.764            & 29.840/0.881              & 29.659/0.876             \\
			ECAFormer \cite{56}                  & - 2024                                    & 26.554/0.823            & 26.472/0.824            & 32.241/0.906              & 32.179/0.903             \\
			DVE-Net \cite{48}                    & TMM 2022                                      & 25.908/0.815            & 25.936/0.815            & 30.711/0.907              & 30.879/0.908             \\
			DCI-Net \cite{2}                    & MM 2023                                     & 26.507/0.831            & 26.460/0.829            & 32.201/0.914             & 32.201/0.913             \\
			\hline
			Ours                      & -                                     & \textbf{26.790/0.834}           & \textbf{26.823/0.834}            & \textbf{32.714/0.915}              & \textbf{32.629/0.913} \\ 
			\midrule[1pt]
		\end{tabular}
	}
	\label{tab:testdata}
\end{table*}

\begin{figure*}[t]\centering
	\includegraphics[width=6.5in, height=5in]{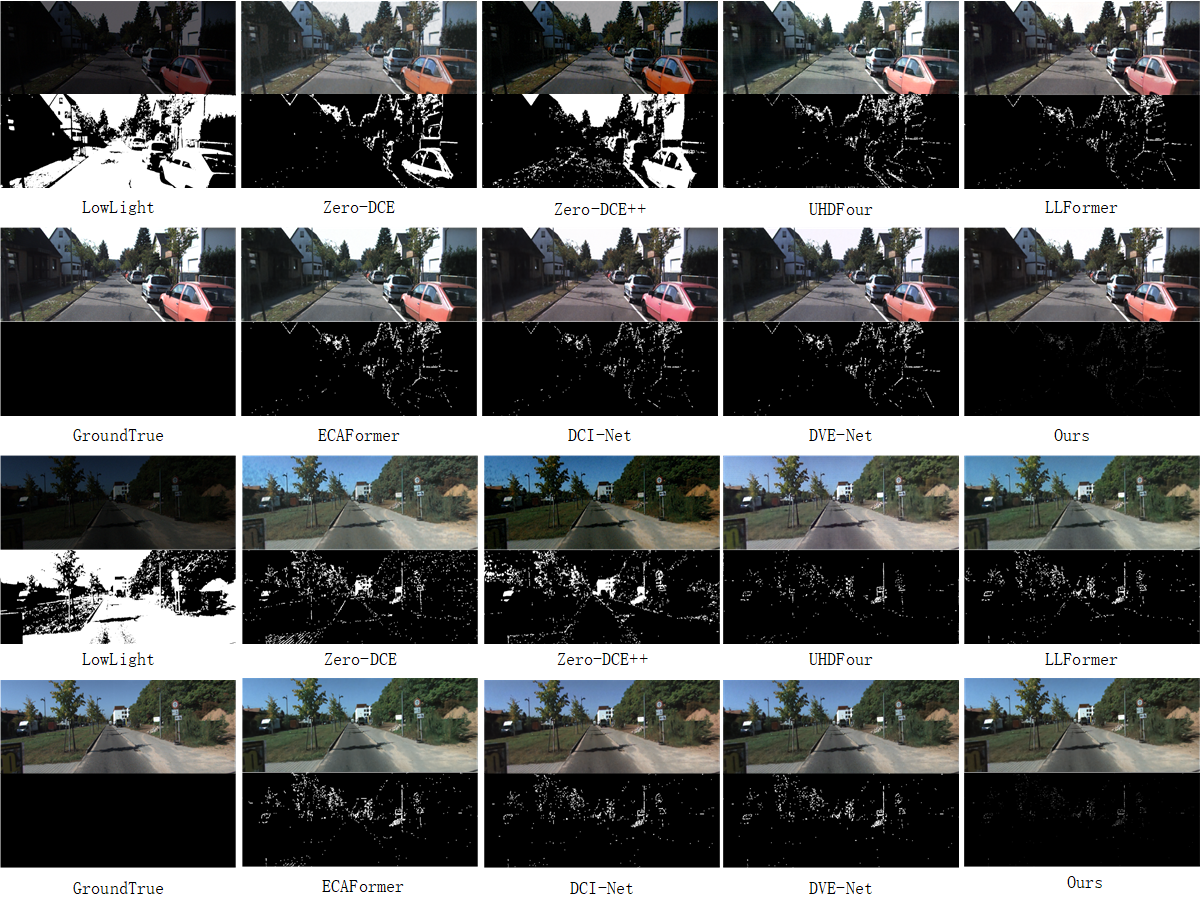}
	\caption{The MSE map between ground truth and the enhanced result on the Kitti2015 dataset, where the white area represents large error and the black area indicate small error. The results show that our method achieves the best enhancement results.}
	\label{pho:kitti}
\end{figure*}
\begin{figure*}[t]\centering
	\includegraphics[width=7in]{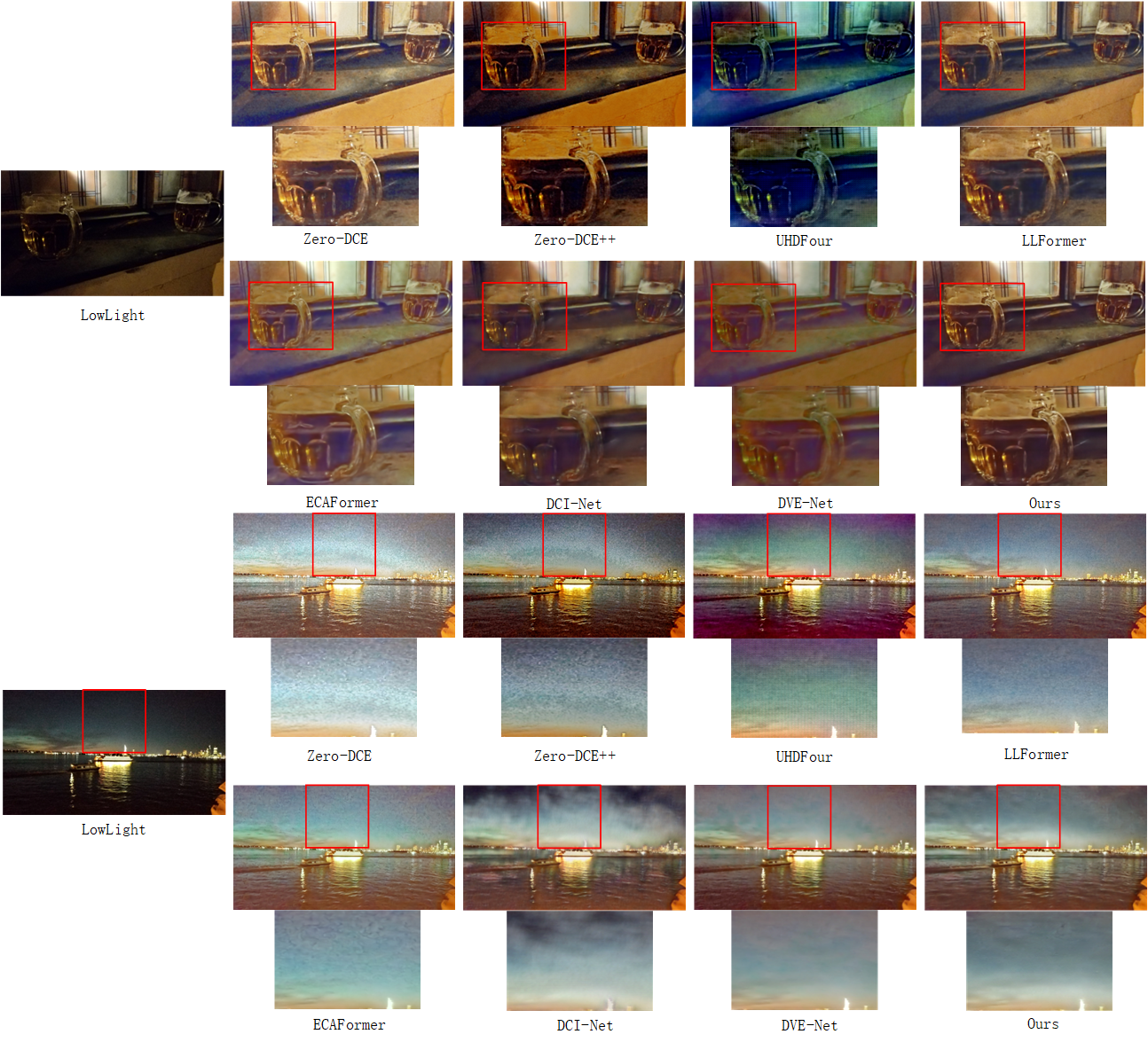}
	\caption{Visualization of enhanced images on the unsupervised real-world dataset Holopix50. Compared with other methods, our approach achieves superior visual results.}
	\label{pho:true}
\end{figure*}

\subsection{Experimental Settings}
\subsubsection{Datasets}
Our training set consists of 1,289 pairs of stereo images, includeing both uniform and non-uniform illumination conditions. Specifically, the uniform illumination subset includes 57 stereo image pairs from Flickr1024, 459 pairs from Holopix50k, and 124 pairs from KITTI. The non-uniform illumination subset comprises 181 stereo image pairs from Flickr1024, 296 stereo image pairs from Holopix50k, and 100 stereo image pairs from KITTI. All images in the training set are synthetic, generated using the data generation method proposed in LFENet \cite{1}. The non-uniform illumination data are used to enhance the model’s robustness and improve its generalization ability in real-world environments.

We use a test set consisting of 312 synthetic stereo image pairs from Flickr1024 and 266 pairs from KITTI to assess the model’s performance. Additionally, we include 191 real-world stereo image pairs from Holopix50k to evaluate the model’s adaptability under complex and realistic lighting conditions. To further facilitate model tuning and performance evaluation, we construct a validation set comprising 30 stereo image pairs, which serves as a reference for monitoring model performance during training.

\subsubsection{Implementation details}
All our experiments are conducted using PyTorch on a machine equipped with an RTX 3090 GPU. We employ the Adam optimizer with a batch size of 20. All training images are randomly cropped to 128×128 pixels. For the validation set, the images are cropped to 400×400, and the batch size is set to 1. The initial learning rate is set to 0.0002 and is halved every 250 epochs. Our network is trained for a total of 1000 epochs.

For evaluation, we adopt two widely-used image quality assessment metrics: PSNR and SSIM. Higher PSNR and SSIM values indicate better quality of the enhanced results. We compare our method with five single-image enhancement techniques (Zero-DCE\cite{46}, Zero-DCE++\cite{57} , UHDFour\cite{10}, LLFormer\cite{47} and ECAFormer\cite{56}) as well as two stereo image enhancement methods (DVENet\cite{48} and DCI Net\cite{2}). All methods are trained on the same dataset as ours and use the hyper-parameters provided in their original papers. They are implemented within the PyTorch framework and run on the same GPU.

\subsection{Quantitative Evaluations}
We conduct tests on two different datasets: Flickr2014 and KITTI2015. Table \ref{tab:testdata} shows the PSNR and Structural Similarity Index Measure (SSIM) values of the images enhanced by our method and the comparison methods. From the table, it can be observed that the unsupervised enhancement methods Zero-DCE and Zero-DCE++ do not perform well, as does the Fourier decomposition-based method UHDFour. The primary reasons for Zero-DCE and Zero-DCE++ is that they only consider illumination adjustment without taking noise into account. UHDFour performs enhancement at a low resolution followed by super-resolution. Super-resolution itself is a challenging problem, and the process often leads to loss of details. In contrast, stereo image enhancement methods achieve better results. Among them, our method outperforms others in terms of both PSNR and SSIM, demonstrating its superior image quality and structural improvement.

Table \ref{tab:NIQE} presents the NIQE scores evaluated on the real-world Holopix50k dataset using a no-reference quality metric. Since lower NIQE scores indicate better image quality, it can be observed that our method achieves the best average NIQE across both left and right views. Compared with other methods, our approach demonstrates competitive overall performance and exhibits strong generalization capability.

\begin{table}[t!]
	\small
	\def\textBF#1{\sbox\CBox{#1}\resizebox{\wd\CBox}{\ht\CBox}{\textbf{#1}}}
	\centering
	\caption{Comparison of NIQE scores for different methods on the real-world Holopix50k dataset.}
	\begin{tablenotes}
		\item
	\end{tablenotes}
	\resizebox{3in}{!}{
	\begin{tabular}{ccccc}
		\midrule[1pt] 
		\multirow{2}{*}{Module} & \multicolumn{2}{c}{Holopix50k (NIQE)}                                         \\ \cline{2-3} 
		& \multicolumn{1}{c}{left} & \multicolumn{1}{c}{right}  \\ \hline
		Zero-DCE \cite{46}           & 3.8012             & 3.827                       \\ \hline
		Zero-DCE++ \cite{57}          & 3.8014              & 3.6950                        \\ \hline
		UHDFour \cite{10}                & 4.6104                     & 4.5659             \\\hline
		LLFormer \cite{47}                 & 3.9516                      & 3.8265             \\ \hline
		ECAFormer \cite{56}                 & 3.648             & 3.815                    \\\hline
		DVE-Net \cite{48}                 & 3.3516             & 3.3265                     \\\hline
		DCI-Net \cite{2}                & 3.4014             & 3.3940                  \\ \hline
		Ours                    & \textbf{3.2623}            & \textbf{3.2383}           \\
		\midrule[1pt]
	\end{tabular}
	}
	\label{tab:NIQE}
\end{table}

\begin{table}[t!]
	\small
	\newsavebox\CBox
	\def\textBF#1{\sbox\CBox{#1}\resizebox{\wd\CBox}{\ht\CBox}{\textbf{#1}}}
	\centering
	\caption{The table displays the the results of the ablation experiments. It can be found that any structure missing will lead to a decrease in performance and cannot achieve optimal results. In the table, W/o represent without the module. }
	\begin{tablenotes}
		\item
	\end{tablenotes}
	\begin{tabular}{ccccc}
		\midrule[1pt] 
		\multirow{2}{*}{Module} & \multicolumn{2}{c}{Left}                            & \multicolumn{2}{c}{Right}                           \\ \cline{2-5} 
		& \multicolumn{1}{c}{PSNR} & \multicolumn{1}{c}{SSIM} & \multicolumn{1}{c}{PSNR} & \multicolumn{1}{c}{SSIM} \\ \hline
		W/o DTEM            & 26.481             & 0.827             & 26.494             & 0.831             \\ \hline
		W/o HF-CIM           & 26.231             & 0.792             & 26.251            & 0.792             \\ \hline
		W/o \(\mathcal{L}_{spa} \)                 & 25.760             & 0.783             & 25.597             & 0.824             \\
		W/o \(\mathcal{L}_{fre} \)                 & 25.447             & 0.815             & 25.270             & 0.831             \\ 
		W/o \(\mathcal{L}_\mathrm{fre}^{(1/8)}\)                 & 25.648             & 0.815             & 25.773             & 0.814             \\
		W/o \(\mathcal{L}_\mathrm{spa}^{(1/8)}\)                 & 25.264             & 0.803             & 25.356             & 0.783             \\
		W/o \(\mathcal{L}_\mathrm{vgg}^{(1/8)}\)                 & 25.426             & 0.793             & 25.435             & 0.827             \\ \hline
		W/o IAM                 & 25.367             & 0.816             & 25.197             & 0.821             \\ \hline
		W/o $F_{\downarrow\times x}(x=2,4,8)$                 & 26.196             & 0.825             & 26.235             & 0.825             \\ \hline
		Ours                    & \textbf{26.790}            & \textbf{0.834}             & \textbf{26.823}             & \textbf{0.834} \\
		\midrule[1pt]
	\end{tabular}
	\label{tab:ablation}
\end{table}

\subsection{Qualitative Evaluation}
Figure  \ref{pho:flickr} display the enhanced images from the Flickr2014 dataset. It can be observed that our method performs well in color and structure restoration compared to other methods. Although some single-view image enhancement methods perform equally well in color correction, there is still a gap in detail recovery compared to stereo image enhancement methods. Compared with other stereo image enhancement methods, our results exhibit fewer pseudo-shadows, better color representation, and are closer to the ground truth.

Figure \ref{pho:kitti} shows the enhanced images and the Mean Squared Error (MSE) maps between the enhanced results and the ground truth on the Kitti2015 dataset, which are converted into binary images. The black regions indicate smaller errors compared to the ground truth, while the white regions indicate larger errors. It can be seen that our method achieves the lowest MSE error compared to the ground truth.

Figure \ref{pho:true} presents the enhanced results on the real unsupervised Holopix50 dataset. We found that most algorithms achieve good results in color restoration, but there is still room for improvement in detail recovery. Our algorithm not only excels in color restoration but also maintains good detail recovery, demonstrating its robustness and generalization ability to adapt to the restoration of real datasets.
\subsection{Ablation Studies}

We present the results of the ablation studies in Table \ref{tab:ablation}.

(1) Impact of DTEM on Performance: Table II shows that if further refinement and recovery of features are not performed after high-frequency feature interaction, the enhancement performance decreases. This indicates that DTEM can enhance the ability to capture details and improve the network's denoising capability.

(2) Impact of HF-CIM on Performance: Table II displays the results on the Flickr2014 dataset. If HF-CIM is missing, the network cannot capture features from the other view. We can see that removing the HF-CIM module leads to a performance drop, both in terms of PSNR and SSIM.

(3) Impact of Each Loss Function on Performance: The results show that eliminating any of the loss functions causes a decline in network performance. By combining these five losses, the quality of the reconstructed image is improved.

(4) Impact of IAM on Performance: IAM includes two stages: spatial information mining and channel information mining. It can be observed that removing either stage results in a performance degradation.

(5) Impact of Downsampling Features $F_{\downarrow\times x}(x=2,4,8)$ on Performance: Table II shows that excluding the fusion of low-frequency information and downsampled features during the encoding stage—using only the low-frequency components obtained from DWT—leads to performance degradation. This indicates that downsampled features can enhance the representation capability of features, thereby contributing to the overall performance improvement.

\section{Conclusion and Future Work}

In this paper, we propose a decoupled approach for low-light stereo image enhancement, in which illumination adjustment and detail restoration are handled separately. Specifically, leveraging the capability of wavelet transform to more effectively separate texture and illumination information, we design two independent branches to process low-frequency and high-frequency information, respectively. The low-frequency branch is responsible for illumination adjustment, while the high-frequency branch focuses on detail restoration and noise suppression. To this end, we introduce a HF-CIM module to perform cross-view and cross-scale interaction on high-frequency components. In addition, we propose a DTEM module to further enhance and denoise the interacted high-frequency features. Extensive experimental results demonstrate that our proposed WDCI-Net achieves state-of-the-art performance compared to other methods and delivers superior visual quality in both illumination correction and texture detail restoration.

The current method employs wavelet transform to decouple low- and high-frequency branches, but it still faces limitations in multi-scale processing efficiency and enhancement under extreme low-light conditions. In future work, we will explore more efficient wavelet variants (e.g., adaptive wavelets) and dynamically optimize the decoupling strategy to improve the collaboration between low-frequency illumination and high-frequency texture, thereby enhancing adaptability to complex low-light scenarios.

\section{Acknowledgments}
This research was funded by National Key Technology R\&D Program of China, grant number 2017YFB1402103-3, Natural Science Foundation of Shaanxi Province, China(2024JC-ZDXM-35,2024JC-YBMS-458), National Natural Science Foundation of China (No.52275511) and Sichuan Science and Technology Program (2023YFG0117), Young Talent Fund of Association for Science and Technology in Shaanxi, China(20240146).
\bibliographystyle{IEEEtran}
\bibliography{reference.bib}

\vspace{-12mm}

	\vspace{-12mm}

\vspace{-12mm}

\vspace{-10mm}

\vspace{-10mm}

\vspace{-170mm}

\end{document}